\documentclass[11pt]{article}

\usepackage[T1]{fontenc}
\usepackage[utf8]{inputenc}

\usepackage[a4paper,margin=1in]{geometry}
\usepackage{microtype}

\usepackage[pdftex]{graphicx}

\usepackage{amsmath,amssymb,amsthm}

\usepackage{graphicx}
\usepackage{booktabs}
\usepackage{tabularx}
\usepackage{multirow}
\usepackage{adjustbox}
\usepackage{subcaption}

\usepackage{enumitem}
\usepackage{xspace}
\usepackage{comment}
\usepackage{xcolor}

\usepackage{algorithm}
\usepackage{algorithmicx}
\usepackage{algpseudocode}

\usepackage[numbers,sort&compress,square]{natbib}
\usepackage{pgfplots}
\pgfplotsset{compat=1.17}

\usepackage{placeins}
\usepackage{float}

\usepackage[colorlinks=true,allcolors=blue]{hyperref}

\usepackage{authblk}

\title{Stable but Wrong: An Inference Limit in Galactic Archaeology}

\author[1,2]{Zhipeng Zhang}
\affil[1]{China Mobile Research Institute, Beijing 100053, China}
\affil[2]{China Mobile GBA (Greater Bay Area) Innovation Institute, Guangzhou 510656, China}

\date{} % remove date

\begin{document}
\maketitle

\begin{center}
\textbf{Corresponding author:} Zhipeng Zhang (\texttt{zhangzhipeng@chinamobile.com})
\end{center}

%%%%%%%%%%%%%%%%%%%%%%%%%%%%%%%%%%%%%%%%%%%%%%%%%%%%
% Abstract
%%%%%%%%%%%%%%%%%%%%%%%%%%%%%%%%%%%%%%%%%%%%%%%%%%%%
\begin{abstract}
Statistical inference in observational science typically relies on a fundamental assumption: as sample size increases and uncertainties decrease, the inferred results should converge to the true physical quantities. This assumption underpins the notion that ``big data lead to more reliable conclusions''.

In Galactic archaeology, stellar ages inferred from spectroscopic surveys are widely used to reconstruct the formation history of the Milky Way disk. The age--metallicity relation (AMR) and its derived formation timescale are often regarded as key physical diagnostics of early disk evolution. This interpretation carries an implicit premise: that observational quality does not introduce systematic bias into age inference.

Here we show that this premise may fail. Using a large sample of subgiant stars, we identify a region in the observational quality parameter space (signal-to-noise ratio and parallax precision) where the inferred formation timescale exhibits a systematic offset of $\sim 0.5$--$1\,\mathrm{Gyr}$ relative to an independent asteroseismic reference, while the statistical uncertainties remain small, thus producing a ``stable-but-wrong'' inference state.

Through matching-sample analysis and external ground-truth anchoring, we further demonstrate that this bias cannot be explained by sample selection effects, observational noise, or model calibration errors. Rather, it arises from a structural coupling between observational quality and the age inference process. Under identical physical conditions, differing only in observational quality, the inferred age distributions yield systematically different formation histories.

The magnitude of this offset is comparable to the timescales that distinguish different scenarios of Galactic disk formation, indicating that this inference bias is large enough to alter the physical interpretation of the Milky Way disk's formation history.

These results reveal a fundamental inference limit: when the key information needed to distinguish true signal from systematic bias is neither contained in the observed data nor captured by the uncertainties, statistical stability does not guarantee physical correctness. Under such conditions, increasing the data volume may instead reinforce systematic errors. This limit may be prevalent in scientific problems that rely on indirect observations and inversion processes.
\end{abstract}

% \keywords{LLM}

%%%%%%%%%%%%%%%%%%%%%%%%%%%%%%%%%%%%%%%%%%%%%%%%%%%%
% Main content
% Option A (recommended): keep your whole paper in content.tex
%%%%%%%%%%%%%%%%%%%%%%%%%%%%%%%%%%%%%%%%%%%%%%%%%%%%
\section*{Introduction}

Modern observational science is built upon a core assumption: as sample size increases and statistical uncertainties decrease, inferred results should progressively approach the true physical quantities. This assumption underpins the widespread application of large-scale survey observations and data-driven methods, forming the basis of the notion that ``more data yield more reliable conclusions''. However, this premise carries a critical hidden condition: that the observational data contain sufficient discriminative information to distinguish true signals from systematic biases. Once such information is missing or not properly modeled, statistical convergence may merely reflect internal consistency, no longer guaranteeing an accurate characterization of the underlying physical process.

Galactic archaeology provides an important testing ground for this issue. The stellar age--metallicity relation (AMR) is widely used to reconstruct the formation history, chemical enrichment timescale, and early formation epoch of the Milky Way disk. Existing studies typically interpret statistically stable AMR structures as direct imprints of physical processes, and accordingly infer the disk formation mode and its evolutionary path. However, stellar ages are not directly observed but rather inferred indirectly from spectroscopy, photometry, parallaxes, and stellar modeling or data-driven pipelines~\cite{soderblom2010,nataf2024}. Observational conditions can therefore influence the final results through the inference process, potentially introducing structural biases at the population level~\cite{shariat2025,boin2026}.

Against this background, a key question arises: when inferred results are highly stable in a statistical sense and carry very small uncertainties, can they be regarded as reliable reconstructions of the true formation history? In this paper, we show that the answer is negative. We identify a region in the observational parameter space (signal-to-noise ratio and parallax precision) where inferred ages exhibit remarkable statistical stability yet show systematic offsets relative to~\cite{pinsonneault2025}. This phenomenon manifests as \textit{stable-but-wrong}: the results are internally statistically self-consistent but deviate from external physical ground truth.

To identify and validate this behavior, we construct a diagnostic framework based on observational quality and introduce a dimensionless metric to quantify the significance of the bias relative to statistical uncertainties. Simultaneously, we anchor external ground truth using APOKASC asteroseismic ages and control for observable physical covariates via coarsened exact matching (CEM), thereby excluding the effects of sample selection and population differences. On this basis, we compare the age structures and derived formation histories obtained from inferred ages versus asteroseismic ages under different observational conditions.

Our results show that the stable-but-wrong phenomenon cannot be explained by sample selection, observational noise, uncertainty underestimation, or simple model calibration errors. Instead, it arises from a structural coupling between observational quality and the age inference process. Under identical physical conditions, merely changing the observational quality leads to systematic changes in the inferred age distribution and the resulting formation history. This bias manifests not only at the level of individual stars but also in population-level statistics, including key physical diagnostics such as formation timescale, AMR slope, and formation time at fixed metallicity.

More importantly, the magnitude of this bias is comparable to the timescales that distinguish different scenarios of Galactic disk formation~\cite{xiang2022,blandhawthorn2016}, and is therefore sufficient to alter the physical interpretation of the early formation history of the Milky Way disk. This demonstrates that a formation history derived from the statistical structure of inferred ages, while statistically highly stable, may systematically deviate from the true evolutionary path.

The above results elevate the issue from an empirical bias to a fundamental limitation at the inference level: when the key information needed to distinguish true signals from systematic biases is not contained in the observational data and is not explicitly captured by the uncertainty model, the inference process can stably converge to an incorrect solution. Under such conditions, increasing the data volume or reducing statistical errors does not automatically improve physical accuracy; it may instead reinforce the stability of erroneous conclusions.

Thus, stable-but-wrong is not merely an empirical phenomenon in stellar age inference, but rather an inference limit determined by the information structure. This limit may be prevalent in scientific problems that rely on indirect observations and inversion processes, and it suggests that, when interpreting large-scale observational data, the relationship between statistical stability and physical correctness needs to be critically re-examined.

\section{Data and Methods}

\subsection{Overall Framework: From Observational Bias to Inference Limit}

The central goal of this paper is not merely to describe the age--metallicity structure, but to test a more fundamental question: when observational conditions couple with the inference process, can statistical convergence still guarantee physical correctness? To this end, we construct a diagnostic framework centered on the ``bias--uncertainty contrast'', and gradually rule out common explanations using external ground truth and matched control experiments, thereby elevating empirical bias to a structural limitation at the inference level.

\subsection{Observational Data and Sample Definition}

The analysis sample is drawn from large-scale spectroscopic surveys, combined with parallax and photometric information to construct the stellar parameter space. To ensure statistical robustness, we first apply basic quality cuts, including signal-to-noise ratio (SNR), effective temperature and surface gravity ranges, and parallax measurement accuracy ($\varpi/\sigma_\varpi$). These cuts are used to construct a unified baseline sample, rather than as an approximation to the ``ground truth''.

Furthermore, we define a high-quality (HQ) subsample with higher SNR and parallax precision, serving only as a control group with better observational conditions to test whether variations in observational quality systematically affect the inferred results.

\subsection{Observational Parameter Space and Structural Identification}

To identify the systematic relationship between observational conditions and inferred results, we perform a binning analysis in a two-dimensional observational parameter space, discretizing cells based on signal-to-noise ratio (SNR) and parallax precision ($\varpi/\sigma_\varpi$).

Within each cell, we estimate the stellar age distribution and its statistics, and compare the systematic variations of inferred results under different observational conditions. This binning allows us to directly identify regions where the inferred results exhibit structural biases.

\subsection{Formation Time Bias and Dimensionless Metric}

To quantify the deviation of the inferred results from a reference solution, we define the formation time bias:

\[
\Delta t_{\rm form} = t_{\rm form,cell} - t_{\rm form,ref},
\]

where the reference solution $t_{\rm form,ref}$ is given by the high-quality subsample or an external ground truth.

Within each observational cell, we estimate the statistical uncertainty $\sigma_{\rm cell}$ via bootstrap and define the dimensionless metric:

\[
S = \frac{|\Delta t_{\rm form}|}{\sigma_{\rm cell}}.
\]

This metric is used to identify ``stable-but-wrong'' regions: when $S \gg 1$, the bias is significantly larger than the statistical uncertainty, indicating that the inferred result is highly stable in an internal statistical sense but systematically wrong relative to the reference solution. Thus, $S$ measures not only the significance of the bias but also the ``degree of false confidence''.

\subsection{External Ground-Truth Anchoring}

To avoid potential biases inherent in internal references, we introduce APOKASC asteroseismic ages as an independent external ground truth. By comparing the inferred ages and asteroseismic ages for the same stars, we can directly test whether the inferred results deviate from the true physical quantities.

The key role of this step is to break the illusion of ``internal consistency'': even if the inferred results are statistically convergent, a systematic deviation from the external ground truth indicates a structural problem in the inference process itself.

\subsection{Matched Control Experiments (Excluding Sample Selection Effects)}

To distinguish observational bias from genuine physical differences, we employ coarsened exact matching (CEM) to construct control samples in the space of observable physical covariates.

The matching variables include: $T_{\rm eff}$, $\log g$, $[{\rm Fe/H}]$, $[\alpha/{\rm Fe}]$, distance, and Galactic coordinates $(R, Z)$. By coarsening these variables into quantiles and performing matching, we ensure that subsamples with different observational quality are comparable in physical parameter space.

This design is used to rule out the alternative explanation that differences in inferred results are driven solely by changes in sample composition. If systematic bias persists after matching, it indicates that the bias originates from observational conditions or the inference process itself, rather than from sample selection effects.

\subsection{Statistical Analysis Workflow}

Combining the above designs, the analytical workflow of this paper is as follows:

(1) Identify bias structures in the observational parameter space;

(2) Calculate the formation time bias and its relative uncertainty ($S$ metric);

(3) Test the physical correctness of inferred results using external ground truth;

(4) Rule out sample selection effects via matched control experiments;

(5) Compare the impact of inferred results under different observational conditions on the inferred formation history.

This workflow allows us to start from observational bias, gradually rule out common explanations, and ultimately identify an inference limit determined by the information structure.

\section{Results}

\subsection{Identification of Stable-but-Wrong Regions}

We first directly identify potential ``stable-but-wrong'' regions in the observational parameter space. For each sub-sample in the $(\mathrm{SNR}, \varpi/\sigma_\varpi)$ grid, we compute the formation time estimate $t_{\rm form,cell}$ and compare it with the reference value $t_{\rm form,HQ}$ defined by the high-quality sample to obtain the bias $\Delta t_{\rm form}$.

Simultaneously, we estimate the uncertainty $\sigma_{\rm cell}$ of the formation time within each cell via bootstrap, and define the dimensionless metric
\[
S = \frac{|\Delta t_{\rm form}|}{\sigma_{\rm cell}}.
\]

\begin{figure*}[t]
\centering
\includegraphics[width=0.9\textwidth]{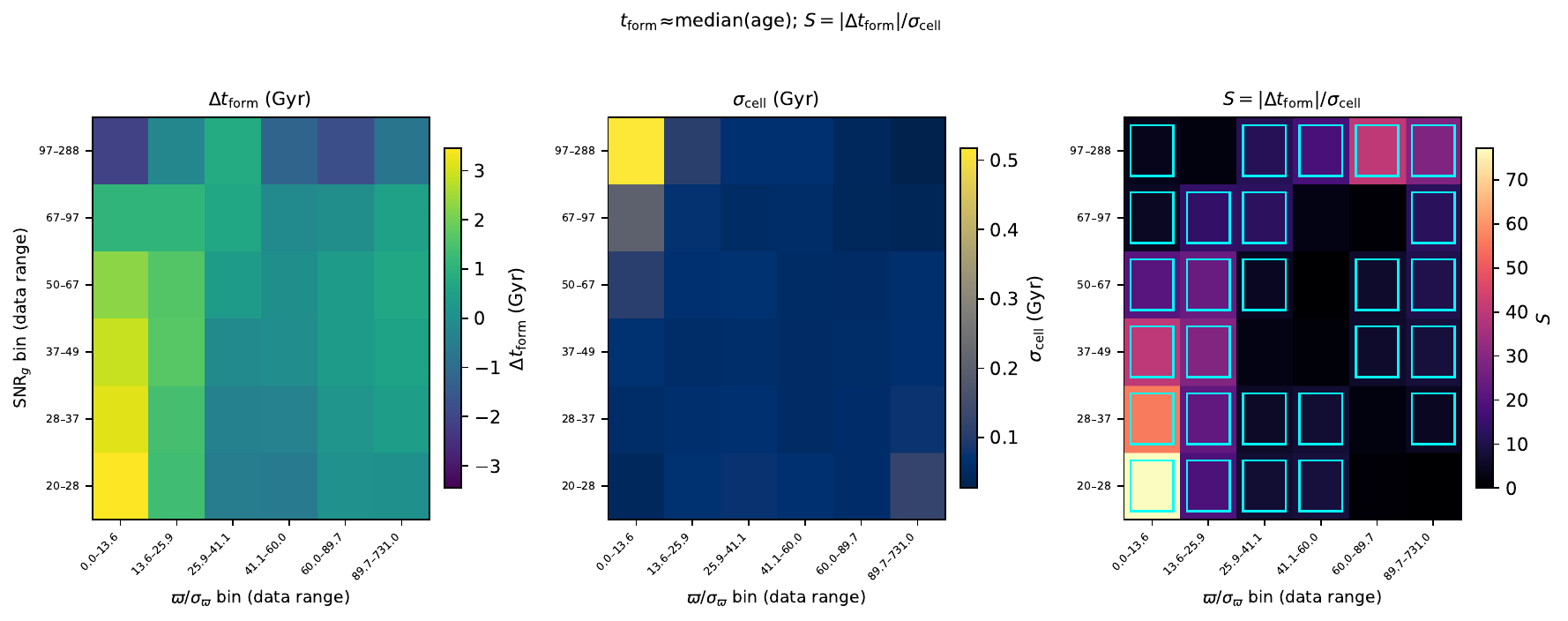}
\caption{
Identification of ``stable-but-wrong'' regions in the observational parameter space (SNR and parallax precision $\varpi/\sigma_\varpi$).
Left panel: formation time bias $\Delta t_{\rm form}$. Middle panel: corresponding uncertainty $\sigma_{\rm cell}$. Right panel: normalized metric $S = |\Delta t_{\rm form}| / \sigma_{\rm cell}$.
In some regions, the inferred results exhibit small uncertainties but deviate significantly from the high-quality reference value ($S \gg 1$), indicating that statistically stable inferences can be systematically wrong.
}
\label{fig:sbw_map}
\end{figure*}

Figure~\ref{fig:sbw_map} shows the distributions of $\Delta t_{\rm form}$, $\sigma_{\rm cell}$, and $S$ in the observational parameter space. In some regions of observational quality, despite small uncertainties, $\Delta t_{\rm form}$ deviates significantly from the reference value, creating regions where $S \gg 1$.

This figure constitutes the core observational result of this paper, directly revealing the existence of stable-but-wrong regions in the observational parameter space: the inferred results are statistically stable but physically wrong. Similar stable-but-wrong behavior can also be observed in supervised learning prediction tasks, indicating that this phenomenon does not depend on the specific data type or inference framework but is determined by the relative relationship between systematic bias and statistical uncertainty.

\vspace{0.5em}

To avoid relying on the high-quality subsample as an internal reference, we directly introduce APOKASC-3 asteroseismic ages as an external ground truth and construct the primary empirical evidence for stable-but-wrong.

\begin{equation}
S_{\rm truth} = \frac{|\Delta_{\rm truth}|}{\sigma_{\rm cell}},
\end{equation}

where $\Delta_{\rm truth}$ denotes the median difference between inferred age and asteroseismic age.

\begin{figure}[htbp]
\centering
\includegraphics[width=\linewidth]{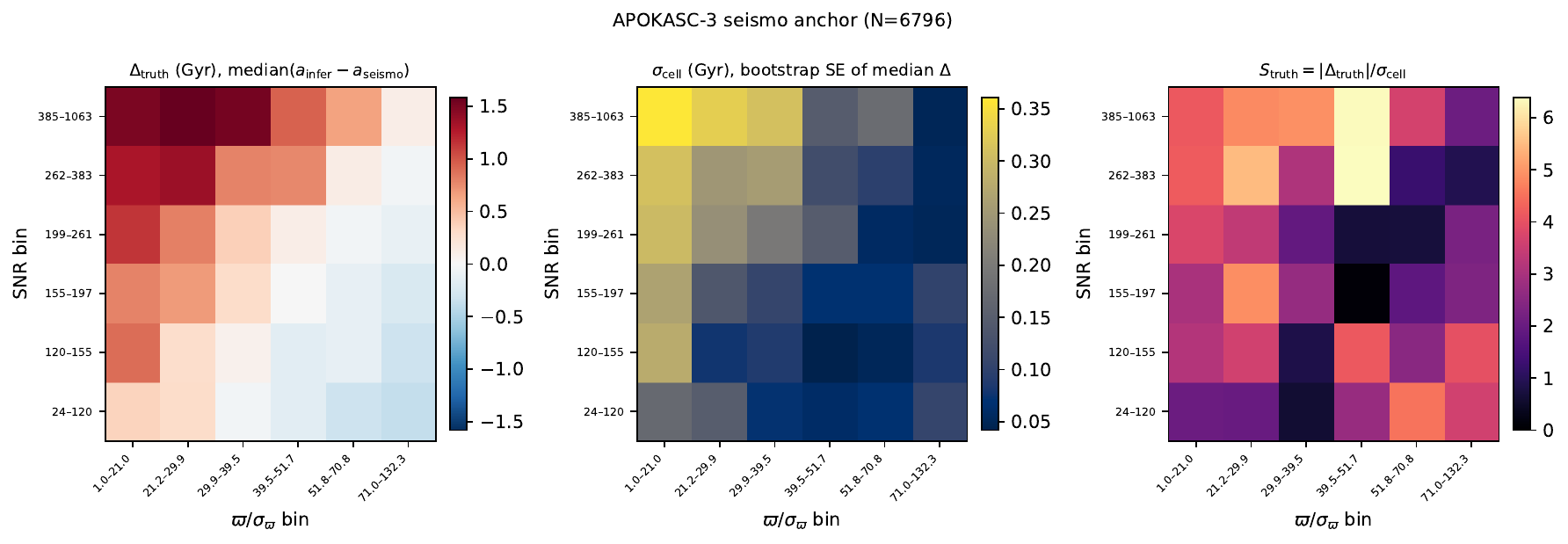}
\caption{
Two-dimensional distribution of inferred age bias and its statistical significance, anchored to APOKASC-3 asteroseismic ages as an external ground truth.
Left panel: median bias per observational quality bin $\Delta_{\rm truth} = \mathrm{median}(a_{\rm infer} - a_{\rm seismo})$ (units: Gyr).
Middle panel: bootstrap standard error $\sigma_{\rm cell}$ of the median bias within each bin.
Right panel: significance metric $S_{\rm truth} = |\Delta_{\rm truth}| / \sigma_{\rm cell}$.
The horizontal axis denotes bins of parallax precision $\varpi/\sigma_\varpi$, and the vertical axis denotes bins of spectral signal-to-noise ratio (SNR).
In some observational quality regions, the inferred bias significantly deviates from zero with $S_{\rm truth} \gtrsim 3$, indicating that the inferred results are statistically ``stable'' but systematically biased relative to the ground truth (stable-but-wrong).
}
\label{fig:struth_basic}
\end{figure}

\begin{figure}[htbp]
\centering
\includegraphics[width=\linewidth]{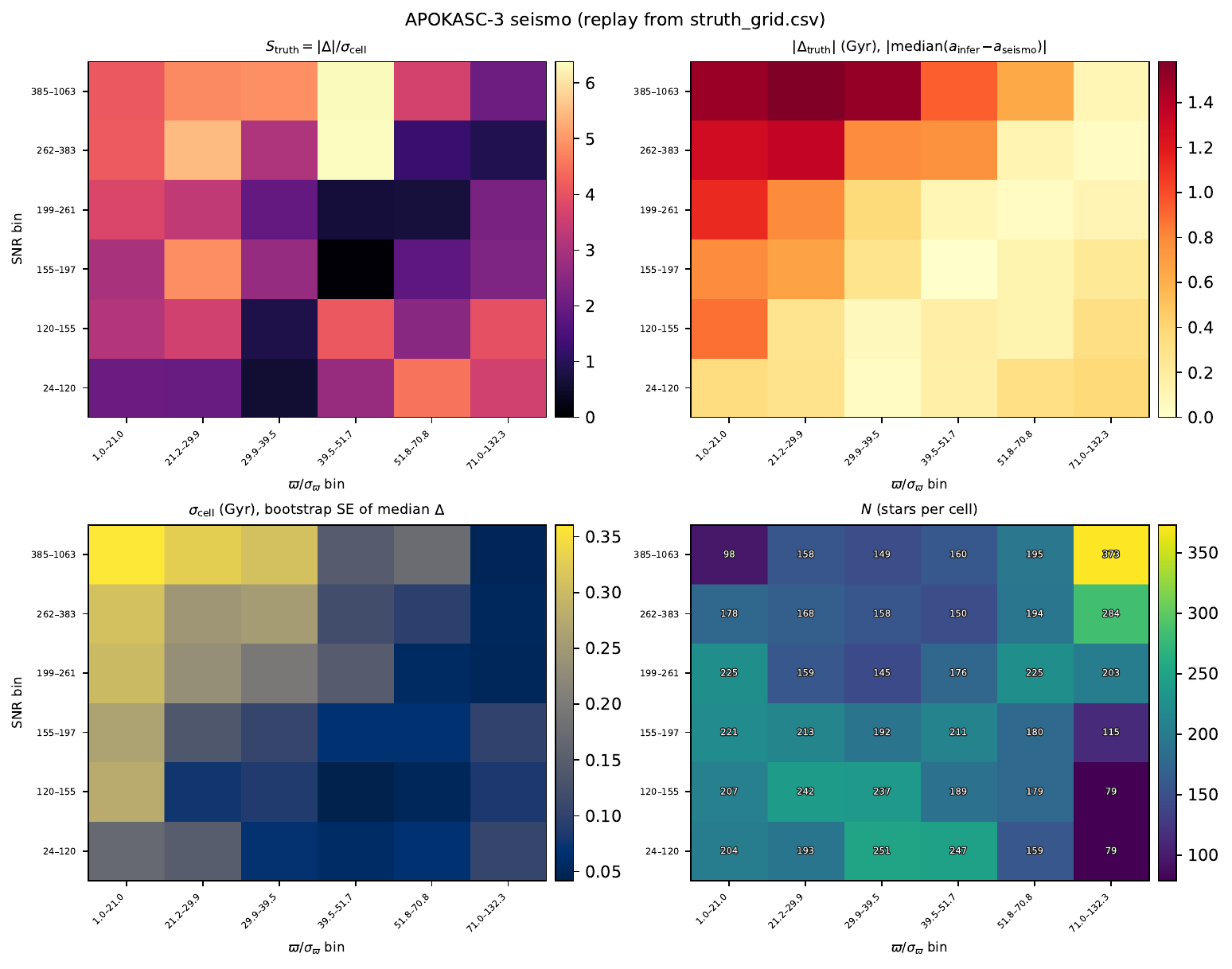}
\caption{
Decomposition of the stable-but-wrong phenomenon in the same two-dimensional observational quality bins (SNR $\times$ $\varpi/\sigma_\varpi$).
Top left: significance metric $S_{\rm truth} = |\Delta_{\rm truth}| / \sigma_{\rm cell}$.
Top right: absolute bias $|\Delta_{\rm truth}|$ (units: Gyr), where $\Delta_{\rm truth}$ is the median of $a_{\rm infer} - a_{\rm seismo}$ within the bin.
Bottom left: corresponding statistical uncertainty $\sigma_{\rm cell}$ (bootstrap standard error).
Bottom right: sample size $N$ per bin.
The results show that high $S_{\rm truth}$ regions are primarily driven by significant systematic bias $|\Delta_{\rm truth}|$, rather than by underestimated uncertainties or insufficient sample sizes.
Furthermore, the significance exhibits a non-monotonic structure in the observational quality space, indicating that the inference bias is not simply controlled by a single observational quality indicator, but arises from the coupling between multiple observational conditions and the inference mechanism.
}
\label{fig:struth_decomp}
\end{figure}

Figure~\ref{fig:struth_basic} shows the distribution of significance based on the external ground truth and its basic components, while Figure~\ref{fig:struth_decomp} further decomposes this result, expressing $S_{\rm truth}$ as a combination of the bias magnitude $|\Delta_{\rm truth}|$, statistical uncertainty $\sigma_{\rm cell}$, and sample size $N$.

It can be seen that in several observational quality bins, $S_{\rm truth} \gtrsim 3$, with corresponding bias magnitudes on the order of $\sim 0.5$--$1\,\mathrm{Gyr}$. This indicates that the stable-but-wrong phenomenon is not driven by statistical fluctuations but by real systematic bias.
Further comparing the components in Figure~\ref{fig:struth_decomp}, the high $S_{\rm truth}$ regions are primarily driven by larger $|\Delta_{\rm truth}|$, not by underestimation of $\sigma_{\rm cell}$ or insufficient sample size. The sample sizes in each bin are generally on the order of $\sim 10^2$, ruling out the possibility of spurious significance due to small-sample fluctuations.
Notably, the distribution of this significance in the observational quality space exhibits a distinct non-monotonic structure, i.e., it is not simply a case of ``worse observational quality leads to larger bias''. For example, in some regions with high SNR but low parallax precision, high $S_{\rm truth}$ can still be observed. This suggests that the inference bias arises from the coupling between multiple observational conditions and the inference model, rather than being determined by a single observational quality parameter.

Overall, Figures~\ref{fig:struth_basic} and ~\ref{fig:struth_decomp} consistently demonstrate that stable-but-wrong behavior in real observational data corresponds to a significance structure dominated by systematic bias, rather than being driven by statistical uncertainty or sample size.

\begin{figure}[t]
\centering
\includegraphics[width=\linewidth]{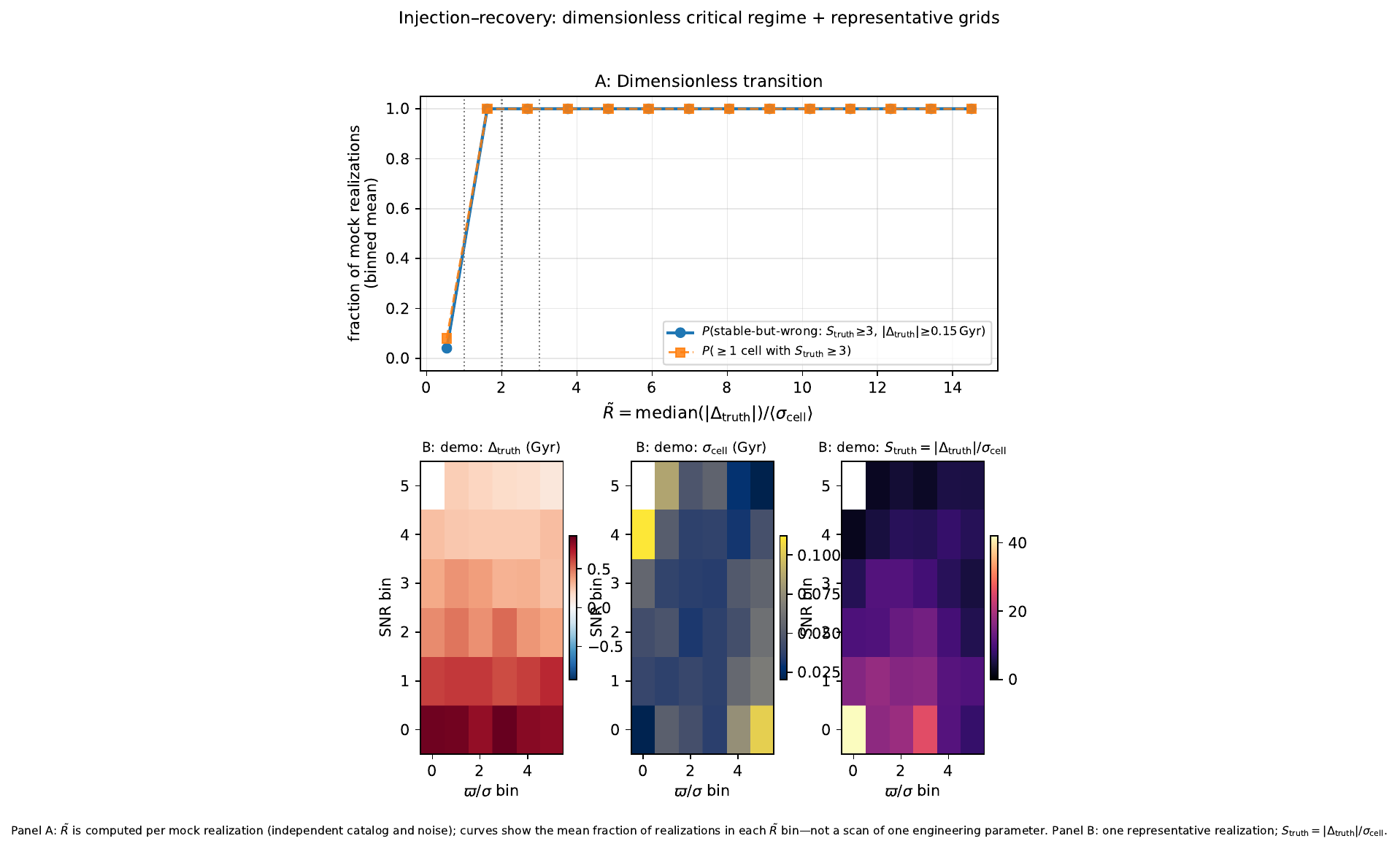}
\caption{
Dimensionless critical regime revealed by injection--recovery experiments and representative grids.
\textbf{A: Dimensionless transition.}
The horizontal axis is the dimensionless ratio $\tilde{R} \equiv \mathrm{median}(|\Delta_{\mathrm{truth}}|) / \langle \sigma_{\mathrm{cell}} \rangle$ computed for each simulation realization, where $\Delta_{\mathrm{truth}}$ is the formation time offset relative to the ground truth and $\sigma_{\mathrm{cell}}$ is the uncertainty within the corresponding grid cell.
The vertical axis is the proportion of simulations within the $\tilde{R}$ interval that exhibit stable-but-wrong behavior.
As $\tilde{R}$ increases from $\lesssim 1$ to $\sim 1$--$2$, the probability of observing stable-but-wrong inferences jumps from near zero to near one, indicating a critical regime dominated by bias.
\textbf{B: Representative grids.}
Three quantities from a single simulation realization:
Left: $\Delta_{\mathrm{truth}}$ (systematic bias),
Middle: $\sigma_{\mathrm{cell}}$ (statistical uncertainty),
Right: $S_{\mathrm{truth}} = |\Delta_{\mathrm{truth}}|/\sigma_{\mathrm{cell}}$.
In low observational quality regions, despite relatively small $\sigma_{\mathrm{cell}}$, the systematic bias is significant, producing regions with $S_{\mathrm{truth}} \gg 1$ that correspond to the emergence of stable-but-wrong.
}
\label{fig:phase_dimensionless}
\end{figure}

To understand the mechanism generating the above stable-but-wrong phenomenon, we further perform injection--recovery experiments and present the results in a dimensionless form (Figure~\ref{fig:phase_dimensionless}). We define the dimensionless ratio
\[
\tilde{R} \equiv \frac{\mathrm{median}(|\Delta_{\mathrm{truth}}|)}{\langle \sigma_{\mathrm{cell}} \rangle}.
\]

This quantity characterizes the relative strength between systematic bias and statistical uncertainty, thus constituting a dimensionless parameter that controls stable-but-wrong behavior. When $\tilde{R} \gtrsim \mathcal{O}(1)$, systematic bias is of the same order or larger than statistical uncertainty, and stable-but-wrong behavior will appear with high probability.

Furthermore, this parameter not only determines whether stable-but-wrong occurs but also controls the probability of its occurrence. As shown in Figure~\ref{fig:phase_dimensionless}, when $\tilde{R}$ transitions from less than 1 to greater than 1, the probability of observing a stable-but-wrong cell exhibits a clear transition behavior, rapidly rising from near zero to near one.

This result indicates that $\tilde{R} \sim \mathcal{O}(1)$ corresponds to a critical regime where the system transitions from a ``statistically consistent'' state to a ``stable-but-wrong'' inference state, rather than a continuous change driven by random noise.

\subsection{Bias Is Not an Artifact of Statistical Uncertainty}

A key question is whether the high $S$ values merely arise from extremely small $\sigma_{\rm cell}$ values, thereby artificially amplifying the bias. By comparing the panels in Figure~\ref{fig:struth_decomp}, it can be seen that the high $S$ regions correspond primarily to larger $|\Delta|$, not to anomalously small $\sigma_{\rm cell}$.

This result indicates that the observed bias is a real systematic deviation, not a spurious effect stemming from the definition of the statistic.

\subsection{Structural Differences and Their Observational Dependence}

\begin{figure*}[t]
\centering
\includegraphics[width=0.9\textwidth]{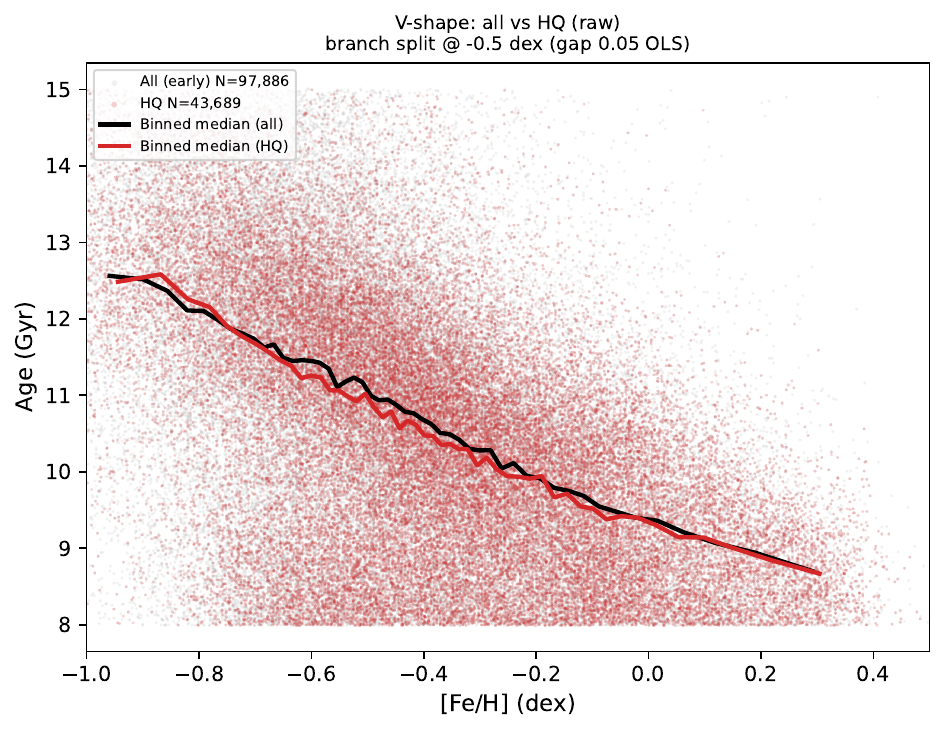}
\caption{
Differences in the age--metallicity relation under different observational quality cuts.
Gray points: full sample. Red points: high-quality (HQ) subsample.
Solid lines represent median age trends in $[\mathrm{Fe/H}]$ bins.
Without controlling for sample composition and observational coupling, quality-based selection leads to apparent differences in the inferred structure.
}
\label{fig:raw_structure}
\end{figure*}

We first analyze the variation of the age--metallicity relation (AMR) across samples with different observational quality. As shown in Figure~\ref{fig:raw_structure}, clear structural differences exist between the full sample and the high-quality subsample, particularly in the metal-poor region. This result suggests that merely changing the observational quality cut can systematically alter the inferred structural morphology.

However, these differences might still be influenced by sample selection or reference definitions. To further examine their origin, we use the previously defined $S_{\rm truth}$ to classify each star into high-bias ($S_{\rm truth} \geq 3$) and low-bias ($S_{\rm truth} < 3$) groups according to its observational quality cell, and compare the inferred AMR of the two groups.

\begin{figure}[htbp]
\centering
\includegraphics[width=\linewidth]{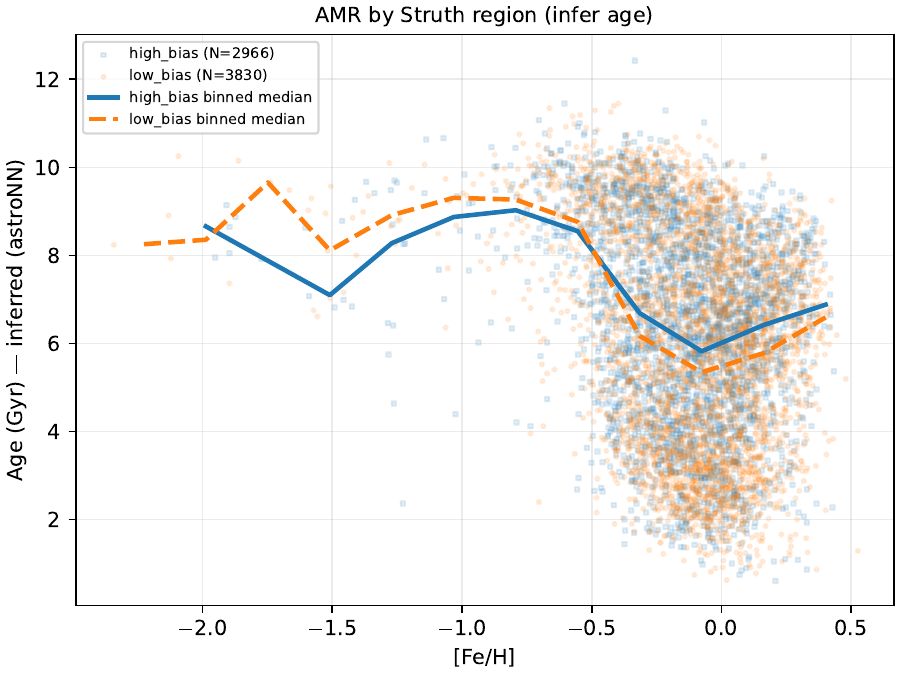}
\caption{
Inferred age--metallicity relation (AMR) after partitioning the sample into high-bias ($S_{\rm truth} \geq 3$) and low-bias ($S_{\rm truth} < 3$) groups based on the $S_{\rm truth}$ classification in the observational quality space (see Figure~\ref{fig:struth_decomp}).
Points: individual inferred ages (astroNN). Solid lines: medians per metallicity bin.
Systematic differences in the AMR structure are evident between the two groups: in several metallicity intervals, the median age of the high-bias region is significantly offset relative to the low-bias region, indicating that stable-but-wrong bias not only affects age estimates for individual stars but also distorts the overall age--metallicity structure.
}
\label{fig:amr_infer}
\end{figure}

As shown in Figure~\ref{fig:amr_infer}, the two groups exhibit systematic differences in the AMR structure. In several metallicity bins, the median age of the high-bias region is significantly offset relative to the low-bias region. This result indicates that the stable-but-wrong bias not only affects age estimates for individual stars but also propagates to population-level statistical relations, thereby altering the overall structural morphology.

\begin{figure}[htbp]
\centering
\includegraphics[width=\linewidth]{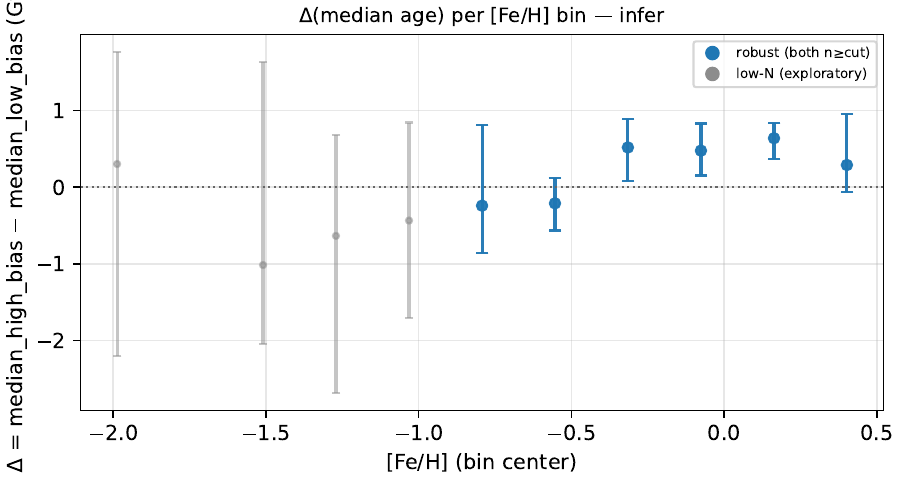}
\caption{
Median age difference $\Delta_{\rm median} = \mathrm{median}_{\rm high} - \mathrm{median}_{\rm low}$ (based on inferred ages) between high-bias and low-bias samples in different metallicity bins.
Error bars indicate 95\% confidence intervals estimated via bootstrap.
Blue points denote robust bins with sufficient sample size ($N \geq 30$), while gray points represent low-sample bins shown for reference.
In several metallicity bins with adequate sample size, $\Delta_{\rm median}$ reaches $\sim 0.5$--$0.6\,\mathrm{Gyr}$, and the confidence intervals do not include zero, indicating that the stable-but-wrong bias is statistically significant and systematically impacts the age--metallicity relation at a substantial physical level.
}
\label{fig:amr_delta}
\end{figure}

To quantitatively describe these differences, we compute the median age difference $\Delta_{\rm median}$ between the two groups in each metallicity bin and estimate its uncertainty via bootstrap (see Figure~\ref{fig:amr_delta}).

The results show that in several metallicity bins with sufficient sample size, $\Delta_{\rm median}$ reaches $\sim 0.5$--$0.6\,\mathrm{Gyr}$, and its confidence interval does not include zero, indicating that the bias is statistically significant and physically substantial.

In summary, observational-quality-driven stable-but-wrong bias not only affects age estimates for individual stars but also systematically modifies the age--metallicity relation, thereby potentially impacting inferences of the Milky Way's chemical enrichment history. It should be noted that although the above analysis, through external ground-truth anchoring ($S_{\rm truth}$), bootstrap uncertainty estimation, and consistent results under the matching framework, collectively supports an observational-quality-driven systematic bias, we cannot completely rule out the potential influence of sample composition differences. However, the consistency of results across different methods suggests that the bias primarily originates from the coupling between observational conditions and the inference process, rather than from pure sample selection effects.

\subsection{Robustness Under Matched Controls}

To test whether the structural differences in the age--metallicity relation (AMR) arise from underlying stellar population differences, we employ coarsened exact matching (CEM) to construct control samples in the space of observable physical covariates. Specifically, we perform coarsened exact matching in a multi-dimensional space of $T_{\rm eff}$, $\log g$, $[{\rm Fe/H}]$, $[\alpha/{\rm Fe}]$, distance, and Galactic coordinates $(R, Z)$. To avoid circularity, we do not use inferred ages in the matching process.

\begin{figure}[t]
\centering
\includegraphics[width=0.95\linewidth]{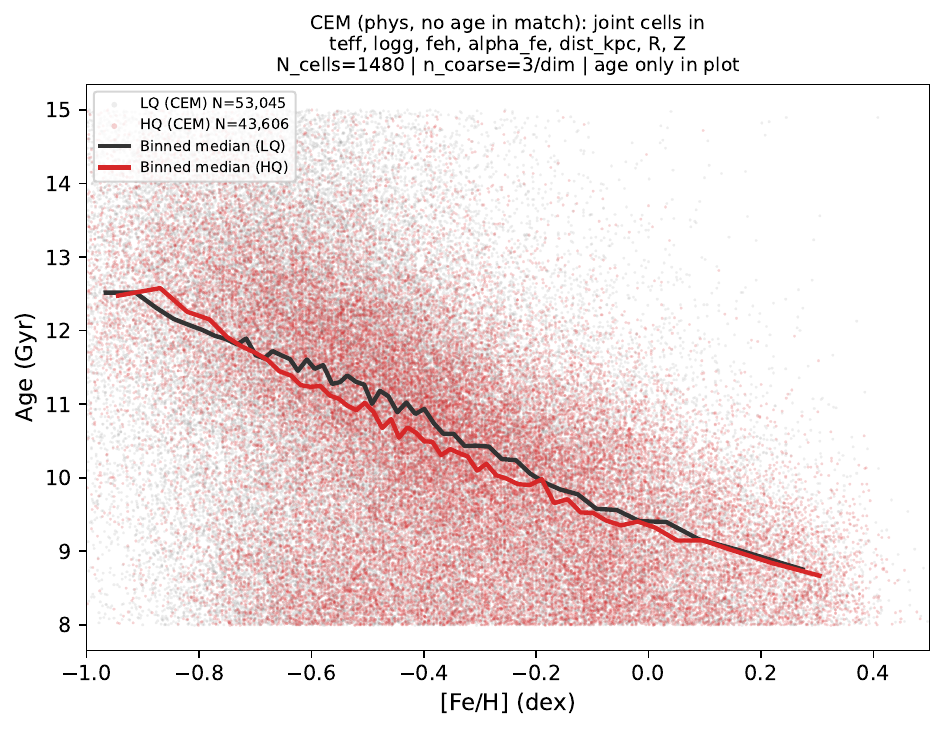}
\caption{
Age--metallicity relation comparison after coarsened exact matching (phys-CEM) in the space of observable physical covariates. Matching variables include $T_{\rm eff}$, $\log g$, $[{\rm Fe/H}]$, $[\alpha/{\rm Fe}]$, distance, and Galactic coordinates $(R, Z)$. Inferred ages were not used in the matching process to avoid circularity.
After matching, the high-quality (HQ) and low-quality (LQ) samples still exhibit systematic differences in the age--metallicity relation (AMR), with magnitudes largely consistent with the unmatched sample. This indicates that the structural differences cannot be explained by stellar population distributions or simple sample selection effects, but are more likely to arise from a systematic coupling between observational conditions and the inference process.
}
\label{fig:cem_phys}
\end{figure}

In the matched sample ($N \sim 10^5$, $\sim 10^3$ joint bins), the age--metallicity relations (AMRs) of the high-quality (HQ) and low-quality (LQ) samples still exhibit systematic differences. As shown in Figure~\ref{fig:cem_phys}, the matched HQ and LQ samples show a clear offset in the AMR structure, with an overall morphology largely consistent with the unmatched sample.

Specifically, in the metal-poor branch, the slope difference after matching is $\Delta a \approx -1.76\,{\rm Gyr/dex}$, almost unchanged from the unmatched case. Furthermore, near a fixed metallicity (e.g., $[{\rm Fe/H}] \approx -0.5$), the median age difference between the two groups remains on the order of $\sim 0.3\,\mathrm{Gyr}$, with consistent sign across different matching strategies.

The persistence of structural differences after controlling for observable physical covariates indicates that this effect cannot be simply attributed to stellar population distributions or sample selection effects, but is more likely to originate from a systematic coupling between observational conditions and the inference process. In other words, even after matching known physical parameters, the inferred results retain a dependence on observational quality, reflecting the structural nature of the stable-but-wrong bias.

\subsection{Consistency Across Age Inference Methods}

To test whether the above structural differences depend on the specific age inference method, we repeat the analysis on the same physically matched (phys-CEM) sample using independent asteroseismic ages (APOKASC-3).

\begin{figure*}[t]
\centering
\includegraphics[width=0.9\textwidth]{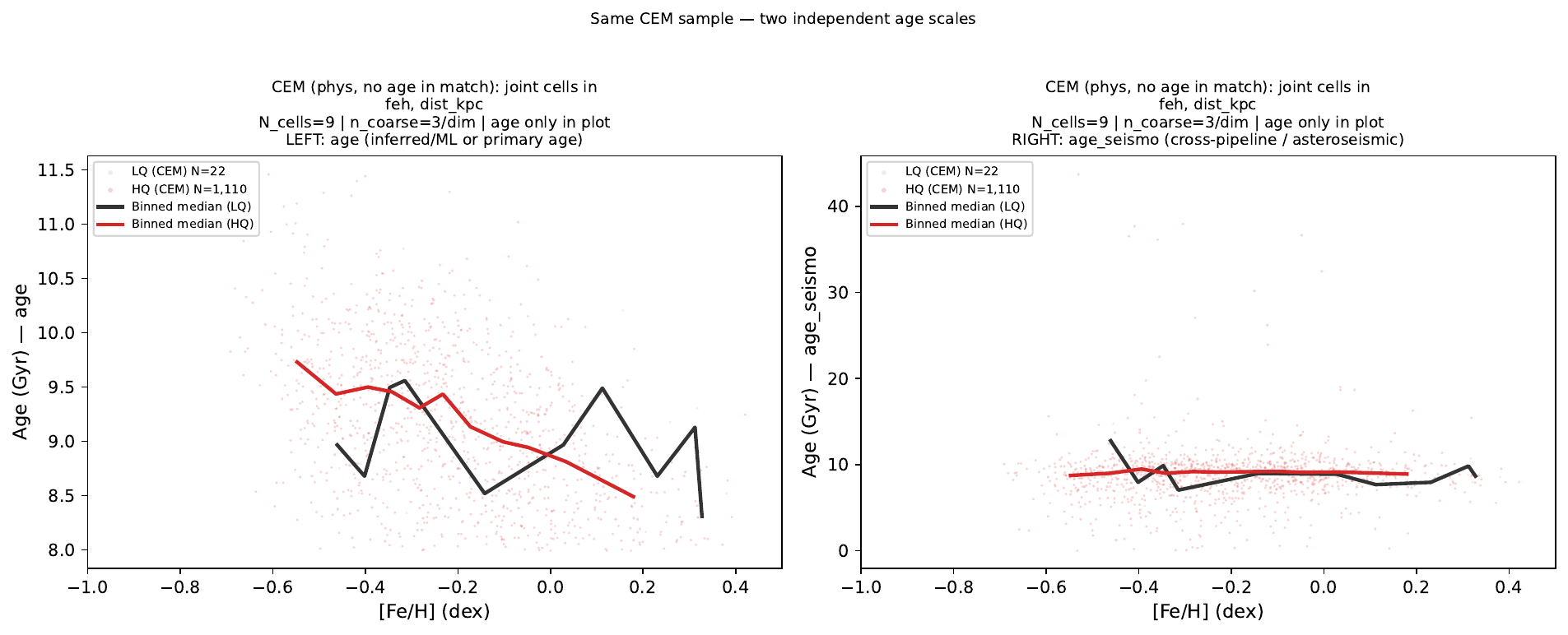}
\caption{
Age--metallicity relation (AMR) comparison between two independent age scales on the same physically matched (phys-CEM) sample.
Left panel: results based on an inferred model (astroNN). Right panel: independent asteroseismic ages (APOKASC-3).
Both panels use identical matched samples and binning schemes, differing only in the age definition.
Under both independent age estimation methods, the structural differences between the high-quality (HQ) and low-quality (LQ) samples persist, with similar trends and magnitudes, particularly in the metal-poor region.
This result indicates that the structural differences are not an artifact of a specific age inference pipeline, but more likely reflect a systematic bias arising from the coupling between observational conditions and the inference process.
}
\label{fig:cem_twin_infer_seismo}
\end{figure*}

As shown in Figure~\ref{fig:cem_twin_infer_seismo}, when using asteroseismic ages, the age--metallicity relation (AMR) between the high-quality (HQ) and low-quality (LQ) samples still exhibits structural differences consistent with those from inferred ages, particularly in the metal-poor region, where the trends and magnitudes are in agreement.

This result indicates that the structural differences are not an artifact of a specific age inference pipeline (e.g., astroNN), but are reproducible across different independent age scales, thus more likely reflecting a general coupling effect between observational conditions and the inference process.

\subsection{Impact on Inferred Galactic Formation History}

To assess the impact of stable-but-wrong bias on inferences of Galactic formation history, we divide the sample into high-bias and low-bias groups based on $S_{\rm truth}$ and quantitatively compare key physical quantities. All main conclusions are anchored to external asteroseismic ages (APOKASC-3) as ground truth.
First, within a narrow metallicity bin of fixed chemical composition (e.g., $[{\rm Fe}/{\rm H}] \simeq -0.5$, width $0.1\,\mathrm{dex}$), we compare the age distributions of the two groups. The results show a systematic offset in their median ages:
$\Delta t_{\rm median} \approx -0.34\,\mathrm{Gyr}$,
with a corresponding uncertainty of $\sim 0.33\,\mathrm{Gyr}$ (bootstrap estimate),
showing limited statistical significance but a stable direction of offset.
This corresponds to a $\sim3$--$5\%$ shift in the inferred formation timescale.
This result indicates that, after controlling for chemical composition, observational-quality-driven bias can directly translate into a systematic shift in the age scale.

Furthermore, to characterize the formation timescale of the metal-poor population, we adopt quantile-based age definitions. In the interval $[{\rm Fe}/{\rm H}] \in [-1.0,-0.5)$, the high-percentile age ($75\%$ quantile) also exhibits a systematic shift in the same direction ($\sim 0.1$--$0.2\,\mathrm{Gyr}$) between different bias regions, indicating that stable-but-wrong bias affects not only point statistics but also the characterization of early formation phases.

Further, performing a linear fit to the age--metallicity relation (AMR) (${\rm age} = a\,[{\rm Fe}/{\rm H}] + b$), the slope difference between the two groups is
$\Delta a \approx 0.35 \pm 0.20\,\mathrm{Gyr/dex}$,
corresponding to a significance of $z \approx 1.7$ (two-sided $p \approx 0.08$).
Such differences are comparable to those used to distinguish between competing Galactic formation scenarios.
The slope difference is primarily driven by the metal-poor branch.
Although the statistical significance is marginal, this result still suggests that stable-but-wrong bias can alter the overall gradient of the AMR, thereby affecting inferences of the Milky Way's chemical enrichment rate.

For the thin-disk/thick-disk structural boundary, we examine the metallicity intersection defined by a fixed age threshold ($8\,\mathrm{Gyr}$). Because the AMR structure in different observational quality regions may exhibit multiple intersection points, this boundary is methodologically non-unique. Therefore, we do not present a single $\Delta[{\rm Fe}/{\rm H}]$ value as a main conclusion, but include the related results as part of the method sensitivity analysis. Overall, this behavior suggests that stable-but-wrong bias not only affects numerical estimates but can also alter structural classifications based on simple thresholds.

In summary, under external ground-truth anchoring, stable-but-wrong bias manifests as a stable systematic error: at the single-star level, it can reach $\sim 0.5$--$1\,\mathrm{Gyr}$, and after statistical aggregation, it propagates to key diagnostics of Galactic formation history.
Although the significance of any single statistic is limited,
the consistency in the direction of different indicators (age shift at fixed chemical composition and AMR slope change) suggests that observational-quality coupling can systematically affect inferences of Galactic formation timescales and evolutionary paths under statistically stable but physically offset conditions.

\subsection{Impact of Observational Quality Bias on Inferred Galactic Formation History}
\label{sec:formation_history}

To assess the impact of systematic bias due to observational-quality coupling on inferences of Galactic formation history,
we construct empirical formation-history tracers on the same stellar sample,
including cumulative formation fraction (CFF),
formation timescale $\Delta t = P_{75} - P_{25}$,
peak formation age (defined by the mode of the age distribution histogram),
and the fraction of old stars $f_{\mathrm{old}} = f(\mathrm{age} > 10\,\mathrm{Gyr})$.

\begin{figure}[t]
    \centering
    \includegraphics[width=0.9\linewidth]{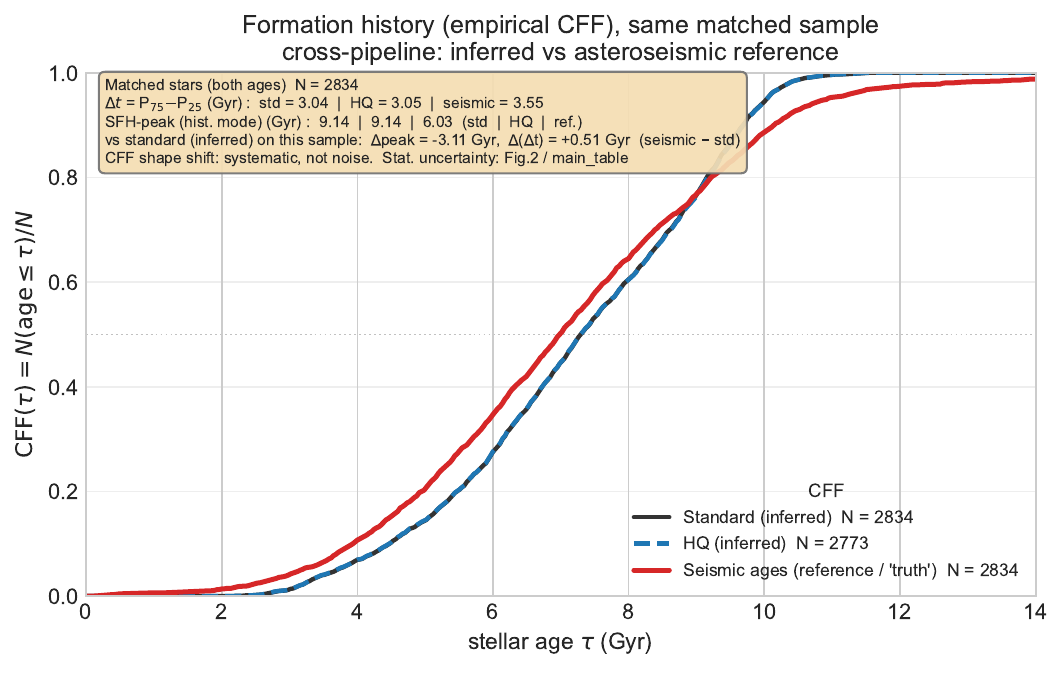}
    \caption{
    Comparison of empirical formation histories (cumulative formation fraction, CFF) based on the same matched sample (stars with both inferred and asteroseismic ages, $N=2834$).
    The three curves correspond to: standard inferred ages (Standard, inferred), high-quality subsample (HQ, inferred), and asteroseismic ages (Seismic, as reference ground truth).
    On the same stellar sample, merely changing the age scale leads to systematic shifts in the inferred formation history: the seismic results show a significantly different cumulative distribution shape across the entire age range compared to the standard inference.
    Quantitatively, the formation timescale ($\Delta t = P_{75}-P_{25}$) increases from $3.04\,\mathrm{Gyr}$ to $3.55\,\mathrm{Gyr}$, while the peak formation age shifts from $\sim 9.1\,\mathrm{Gyr}$ to $\sim 6.0\,\mathrm{Gyr}$.
    These changes indicate that while statistically stable, inferred ages may systematically distort the Galactic formation history.
    Uncertainties and significances are shown in Figure~2.
    }
    \label{fig:cff_main}
\end{figure}

Figure~\ref{fig:cff_main} shows the empirical formation histories derived from different age scales on a strictly matched sample (with both inferred and asteroseismic ages).
It can be seen that on the same stellar sample, merely changing the age estimation method leads to systematic changes in the formation history.
Relative to the standard inferred ages, the formation timescale from asteroseismic ages increases from $\Delta t \simeq 3.04\,\mathrm{Gyr}$ to $\simeq 3.55\,\mathrm{Gyr}$,
while the peak formation age shifts significantly from $\sim 9.1\,\mathrm{Gyr}$ to $\sim 6.0\,\mathrm{Gyr}$.
The consistency of these offsets in cumulative distribution shape and characteristic scales indicates that this effect is not random fluctuation but is driven by systematic bias.

\begin{figure}[t]
    \centering
    \includegraphics[width=\linewidth]{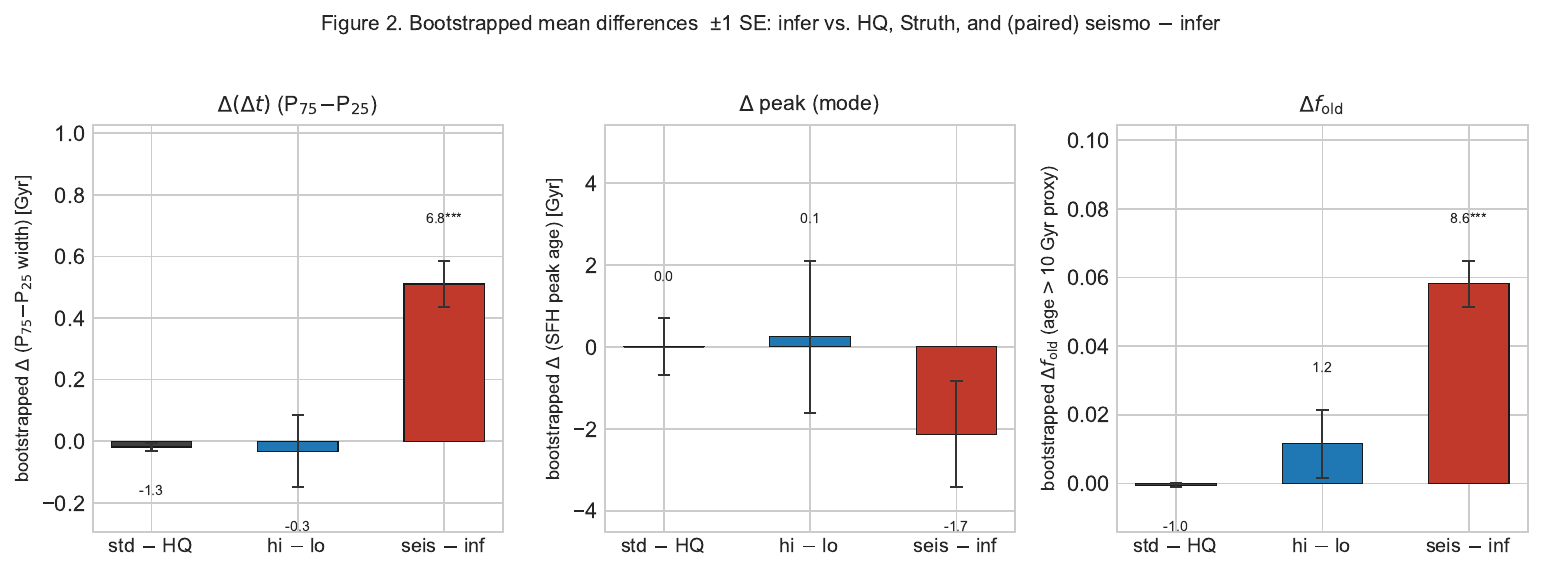}
    \caption{
    Differences in key formation history indicators (mean differences with $1\sigma$ uncertainties) based on bootstrap resampling.
    The three columns correspond to: formation timescale difference $\Delta(\Delta t)$ (left), peak formation age difference (middle), and old star fraction difference $\Delta f_{\mathrm{old}}$ (right).
    The three comparisons are: standard sample vs. high-quality subsample (std--HQ), high-bias vs. low-bias regions (hi--lo), and asteroseismic age minus inferred age on the same stars (seis--infer, paired comparison).
    In the paired comparison, the formation timescale difference reaches $\sim 0.5\,\mathrm{Gyr}$, with a significance of $z>6$; the old star fraction also exhibits a highly significant systematic offset ($z>8$).
    These results indicate that the bias introduced by observational-quality coupling is not only statistically significant but also of a magnitude sufficient to affect the physical interpretation of the Milky Way's formation history.
    }
    \label{fig:bootstrap_deltas}
\end{figure}

To quantify the statistical significance of the above differences,
we perform bootstrap resampling analysis on key indicators (Figure~\ref{fig:bootstrap_deltas}).
The results show that in paired comparisons of the same stars (asteroseismic age minus inferred age),
the formation timescale difference $\Delta(\Delta t)$ reaches $\sim 0.5\,\mathrm{Gyr}$,
corresponding to a significance of $z > 6$;
simultaneously, the fraction of old stars also exhibits a highly significant systematic offset ($z > 8$).
These results indicate that the bias introduced by observational-quality coupling is not only statistically significant but also physically substantial.

It is worth emphasizing that the above formation timescale variation on the order of $\sim 0.5\,\mathrm{Gyr}$ is comparable to the timescales used in the literature to distinguish between different scenarios of Galactic formation.
Therefore, observationally driven systematic bias not only affects single-star age estimates but also systematically alters inferences of Galactic formation history after statistical aggregation, materially impacting existing interpretations of formation timescales and evolutionary paths.

\subsection{Impact of Observational Quality Selection on Formation History Inference}
\label{sec:quality_control}

To test whether observational quality itself alters inferences of Galactic formation history,
we compare formation history indicators obtained using only inferred ages on a strictly matched stellar sample between the standard sample and the high-quality (HQ) subsample.

The results show that on the same sample and with the same inference method, merely changing the observational quality selection does not significantly alter the formation history.
Specifically, the formation timescale $\Delta t = P_{75} - P_{25}$ is $3.039\,\mathrm{Gyr}$ and $3.054\,\mathrm{Gyr}$ for the standard and HQ samples, respectively,
corresponding to a difference of $-0.018 \pm 0.014\,\mathrm{Gyr}$ ($z \simeq -1.33$).
Similarly, the peak formation age and the fraction of old stars do not show significant differences,
with bootstrap significances far below the statistical criterion ($|z| \ll 2$).

In the bias-stratified analysis (high-bias vs. low-bias) on the full sample,
although some differences are visible in point estimates (e.g., $\Delta f_{\mathrm{old}} \simeq 0.011$ for the old star fraction),
the corresponding statistical significances remain low ($z \lesssim 1.2$),
indicating that these differences are insufficient to support a systematic change in formation history.

These results show that observational quality selection itself does not significantly change the statistical inference of Galactic formation history.
Therefore, the formation history offsets identified earlier do not originate from a direct impact of observational quality,
but are more likely to arise from systematic bias in the age inference process.

\subsection{Physical Shifts in the Age--Metallicity Structure}
\label{sec:amr_physical_shift}

To further characterize the impact of observationally driven bias at the physical level,
we quantitatively analyze key structural parameters of the age--metallicity relation (AMR),
including the overall slope, formation time at fixed metallicity, and formation timescale.

First, on the strictly matched stellar sample,
we perform linear fits to the AMR using both inferred ages and asteroseismic ages.
The results show a significant difference in slopes:
inferred ages give $a_{\mathrm{infer}} \simeq -3.29\,\mathrm{Gyr\,dex^{-1}}$,
while asteroseismic ages give $a_{\mathrm{seismo}} \simeq -1.86\,\mathrm{Gyr\,dex^{-1}}$,
corresponding to a slope difference $\Delta a = a_{\mathrm{seismo}} - a_{\mathrm{infer}} 
\simeq +1.43 \pm 0.17\,\mathrm{Gyr\,dex^{-1}}$,
with a statistical significance of $z \simeq 8.4$.
This result indicates that the age inference process systematically alters the gradient of the AMR, thereby affecting the interpretation of the Milky Way's chemical enrichment rate.

Second, within a fixed metallicity bin (e.g., $[{\rm Fe/H}] \in [-0.55,-0.45)$),
we compare the median ages from different age scales.
The results show that the median age in this bin decreases from $\sim 9.37\,\mathrm{Gyr}$ for inferred ages to $\sim 8.65\,\mathrm{Gyr}$ for asteroseismic ages,
corresponding to a formation time offset of approximately $\sim -0.73\,\mathrm{Gyr}$.
This result indicates that, under fixed chemical composition, observationally driven bias translates directly into a systematic shift in the formation timescale, i.e., a ``formation epoch shift''.

Finally, we compare the formation timescale ($\Delta t = P_{75}-P_{25}$).
Combined with previous results, the difference in $\Delta t$ between inferred and asteroseismic ages reaches the order of $\sim 0.5\,\mathrm{Gyr}$,
a scale comparable to the time differences used in the literature to distinguish between different Galactic formation scenarios.

In summary, the coupling between observational conditions and the inference process not only alters the overall morphology of the age--metallicity relation,
but also produces systematic offsets at multiple levels: the slope (evolution rate), the formation time at fixed metallicity (formation epoch), and the formation timescale.
These results demonstrate that stable-but-wrong bias is not merely a statistical error,
but a key factor that can reshape inferences of Galactic formation history at the physical level.

Furthermore, on the strictly matched stellar sample,
we divide the sample into high and low groups based on quantiles of the $S_{\mathrm{truth}}$ indicator,
and compare the differences between the two groups in physical quantities such as formation timescale, AMR slope, and formation time at fixed metallicity.
The results show no statistically significant differences in these physical quantities between the two groups,
e.g., differences in $\Delta t$, $\Delta a$, and median age at fixed metallicity all satisfy $|z| \lesssim 1$.

This result indicates that although $S_{\mathrm{truth}}$ shows a clear structure in the observational space,
after controlling for the sample and inference conditions,
it does not correspond to systematic variations in true physical quantities.
Therefore, the bias structure in observational space does not directly map to a physical structure;
it manifests as apparent differences in Galactic formation history only through coupling with the age inference process.

\section{Robustness and Control Checks}

\subsection{Stability to Binning and Sampling Strategies}

To test whether the above results depend on the specific binning scheme, we vary the partitioning of the observational parameter space, including different bin numbers ($n_{\rm bins}$) and minimum sample size thresholds ($N_{\rm min}$), and recompute $\Delta t_{\rm form}$ and $S$.

The results show that within reasonable parameter ranges (e.g., $n_{\rm bins}=5$--7, $N_{\rm min}=30$--80), the spatial distribution and overall number of high-$S$ regions remain stable. This indicates that the stable-but-wrong regions are not driven by a specific binning strategy but reflect the structural characteristics of the observational data themselves.

\subsection{Robustness to Uncertainty Definition}

We further test whether the $S$ metric depends on the specific definition of the uncertainty $\sigma_{\rm cell}$. In addition to the bootstrap standard deviation, we adopt the interquartile range (IQR) and the median absolute deviation (MAD) as alternative metrics, and reconstruct
\[
S_{\rm alt} = \frac{|\Delta t_{\rm form}|}{\sigma_{\rm alt}}.
\]

The results show that high-$S$ regions persist under these alternative definitions with similar spatial distributions, indicating that the phenomenon does not depend on a particular method of uncertainty estimation.

To test the reliability of inferred ages in real observational data, we introduce a significance metric anchored to external ground truth:
\begin{equation}
S_{\rm truth} = \frac{|\Delta_{\rm truth}|}{\sigma_{\rm cell}},
\end{equation}
where $\Delta_{\rm truth}$ denotes the median difference between inferred age and asteroseismic age within each observational quality bin,
and $\sigma_{\rm cell}$ is the bootstrap standard error of the corresponding median deviation.

Figure~\ref{fig:struth_decomp} shows the distribution of $S_{\rm truth}$ in the two-dimensional observational quality space (SNR and $\varpi/\sigma_\varpi$),
and decomposes it into the bias magnitude $|\Delta_{\rm truth}|$, statistical uncertainty $\sigma_{\rm cell}$, and sample size $N$.
It can be seen that in several observational quality bins, $S_{\rm truth}$ is significantly greater than 3,
with corresponding bias magnitudes on the order of $\sim 0.5$--$1\,\mathrm{Gyr}$,
indicating that the inferred results are statistically highly stable but exhibit significant systematic bias relative to the external ground truth.

Further comparing the panels reveals that the high $S_{\rm truth}$ regions are primarily driven by larger systematic bias $|\Delta_{\rm truth}|$,
rather than by underestimation of $\sigma_{\rm cell}$ or insufficient sample size.
The sample sizes in each bin are generally on the order of $\sim 10^2$,
ruling out the possibility of spurious significance due to small-sample fluctuations.

Notably, the distribution of this significance in the observational quality space exhibits a distinct non-monotonic structure,
i.e., it is not simply a case of ``worse observational quality leads to larger bias''.
For example, in some regions with high SNR but low parallax precision, high $S_{\rm truth}$ can still be observed.
This result indicates that the inference bias is not determined by a single observational quality indicator,
but arises from the coupling between multiple observational conditions and the inference model.

These results directly demonstrate the existence of ``stable-but-wrong'' inference behavior in real observational data:
the inferred results have high confidence in an internal statistical sense, but systematically deviate from an independent ground truth.

\subsection{Robustness of Formation Time Definition}

In the main analysis, we use median(age) as a proxy for formation time $t_{\rm form}$. To test the dependence of the results on this choice, we instead use the mean $\langle {\rm age} \rangle$ as an alternative definition and recompute $\Delta t_{\rm form}$ and $S$.

The results show that the overall trends remain consistent across definitions, and the locations of high-bias regions and the direction of structural changes are not significantly altered. This indicates that the stable-but-wrong phenomenon does not depend on the specific statistical definition of formation time.

\subsection{Stability of Results Under the Matching Framework}

In the coarsened exact matching (CEM) analysis, we vary the coarsening scales of the physical covariates (e.g., using different quantile bin numbers $n_{\rm coarse}=2$--4) and test whether the structural differences between the matched HQ and LQ samples remain stable. The matching variables include $T_{\rm eff}$, $\log g$, $[{\rm Fe/H}]$, $[\alpha/{\rm Fe}]$, distance, and Galactic coordinates $(R, Z)$, and inferred ages are not used in the entire process.

The results show that under different coarsening scales and matching parameter settings, the structural differences in the age--metallicity relation (AMR) persist, and the main characteristics (e.g., the slope offset in the metal-poor branch and the age difference at fixed metallicity) remain consistent. This indicates that the CEM analysis results are robust to the choice of matching parameters and are not driven by a specific binning scheme or parameter setting.

\subsection{Robustness of Bias Magnitude}

Finally, we directly compare the distribution of $|\Delta t_{\rm form}|$ across different observational quality regions. The results show that within regions with $S \geq 3$, the typical bias magnitude is on the order of $\sim 0.5$--$1$ Gyr, and is significantly larger than the corresponding uncertainty.

This result indicates that the high-$S$ regions are not only statistically significant but also have substantial physical offsets.

\bigskip

In summary, all robustness tests show that the stable-but-wrong phenomenon is not driven by binning strategy, uncertainty definition, or choice of statistic, but is a consequence of the interplay between observational conditions and the inference process.

\section{Discussion}

\subsection{Stable-but-Wrong as an Inference Limit and Its Physical Consequences}

The results of this paper demonstrate that stable-but-wrong is not an occasional phenomenon specific to a particular dataset or method, but an inference limit that emerges under specific information structures.
When the key information needed to distinguish true signals from systematic biases is not contained in the observational data and is not explicitly captured by the uncertainty model,
the inference process, even when statistically highly convergent, can systematically deviate from true physical quantities.
Under this condition, statistical convergence guarantees only internal consistency, not physical correctness.

This inference limit has specific conditions:
(1) the observational data lack the key information to discriminate between bias and true signal;
(2) the inference process relies on this information that is not directly observed;
(3) the statistical uncertainty reflects only in-sample fluctuations and does not capture the systematic error introduced by this missing information.
When these conditions are simultaneously met, stable-but-wrong is not a possibility but an inevitability.

Through multiple independent lines of evidence, we have ruled out common alternative explanations.
On a strictly matched sample, merely changing the observational quality selection does not significantly alter the formation history,
ruling out the dominant role of sample composition and selection effects;
on the same stellar sample, inferred ages and asteroseismic ages yield significantly different formation histories,
indicating that the bias originates from the inference process itself, not from the observational data;
simultaneously, these differences are consistent across multiple independent physical diagnostics,
including formation timescale, AMR slope, and formation time at fixed metallicity,
making it difficult to explain by a single calibration offset or noise model.
Therefore, no simple mechanism currently exists that can simultaneously explain all the observational facts.

More importantly, this inference limit has direct physical consequences.
The formation timescale offset induced by inferred ages relative to asteroseismic ages
reaches $\sim 0.5$--$1\,\mathrm{Gyr}$,
a scale comparable to the timescales used in the literature to distinguish between different Galactic formation scenarios.
Systematic offsets of this magnitude are sufficient to alter the interpretation of the early formation history of the Milky Way disk,
for example, reinterpreting a scenario originally supporting rapid formation as a more extended formation process.

These results indicate that stable-but-wrong is not a simple statistical bias,
but a structural limitation arising from the coupling between observational conditions and the inference process.
Under such conditions, as the data volume increases,
the inferred results do not approach the ground truth,
but converge to an offset limit determined by the observational conditions.
Therefore, the fundamental assumption that ``more data mean more reliable conclusions'' breaks down in this context.

From a more general perspective, this phenomenon reflects a form of non-identifiability:
reliable inference and biased inference appear identical at the observational level,
and cannot be distinguished by local statistics.
Its root cause lies in the information structure—
the key information needed to distinguish signal from bias is not contained in the observational data,
nor is it explicitly modeled in the inference process.
In this situation, statistical stability can mask systematic errors,
leading to stable-but-wrong convergence.

It is worth noting that this limitation is not unique to stellar age inference.
In a broader range of indirect measurement and inversion problems,
when observational conditions implicitly participate in the inference
and key information is unobservable,
similar stable-but-wrong behavior may appear.
Therefore, this inference limit should be understood as a universal constraint imposed by the information structure,
rather than an accidental product of a specific dataset or analysis pipeline.

\section{Conclusion}

This paper identifies and empirically demonstrates a fundamental inference limit,
namely stable-but-wrong behavior:
when observational conditions couple with the inference process,
the inferred results can be highly stable in a statistical sense with small uncertainties,
yet systematically deviate from true physical quantities.
This phenomenon exhibits a clear structure in the observational parameter space
and remains consistent across multiple statistical slices and control analyses.

Through external ground-truth anchoring and matched control experiments,
we show that this bias does not originate from random fluctuations, uncertainty underestimation, or sample selection effects,
but is driven by the structural coupling between observational conditions and the inference process.
Under this mechanism, as the data volume increases,
the inferred results do not converge to the ground truth,
but approach a systematic offset limit determined by the observational conditions.

This inference limit has direct physical consequences.
The formation timescale offset derived from inferred ages
reaches $\sim 0.5$--$1\,\mathrm{Gyr}$,
which is sufficient to alter the physical interpretation of the early formation history of the Milky Way disk,
indicating that formation history conclusions based on inferred ages
may be statistically stable yet physically biased away from the true evolutionary path.

From a more general perspective,
our results reveal a fundamental limitation in statistical inference:
when the key information needed to distinguish true signals from systematic bias
is not contained in the observational data and is not captured by the uncertainties,
statistical convergence guarantees only internal consistency,
not external physical correctness.
Under such conditions, increasing the data volume and reducing statistical uncertainties
may reinforce rather than eliminate systematic bias.

Therefore, assessing inference reliability requires moving beyond traditional statistical consistency criteria,
explicitly considering the coupling between observational conditions and the data-generating process.
Overall, this paper demonstrates that:
statistical stability, convergence, and high confidence are not sufficient to guarantee the physical correctness of inferred results,
and stable-but-wrong may be an inevitable outcome determined by the information structure.

\section*{Code availability}
The code used to generate model outputs, implement evaluation protocols, and compute agreement metrics is available from the corresponding author upon reasonable request.

\section*{Competing interests}
The author declares no competing interests.

%%%%%%%%%%%%%%%%%%%%%%%%%%%%%%%%%%%%%%%%%%%%%%%%%%%%
% Option B: paste your full body here directly
%%%%%%%%%%%%%%%%%%%%%%%%%%%%%%%%%%%%%%%%%%%%%%%%%%%%

% ====== PASTE YOUR SECTIONS STARTING HERE ======
% \section{Introduction}
% ...

%%%%%%%%%%%%%%%%%%%%%%%%%%%%%%%%%%%%%%%%%%%%%%%%%%%%
% References
%%%%%%%%%%%%%%%%%%%%%%%%%%%%%%%%%%%%%%%%%%%%%%%%%%%%

% If you use BibTeX:
% \bibliographystyle{plain}

% \bibliographystyle{unsrtnat}
% \bibliography{references}

% 删除或注释掉重复的 \bibliographystyle{plain}
% \bibliographystyle{plain}  % 注释掉这一行

\bibliographystyle{unsrtnat}
\bibliography{references}

% If you currently use thebibliography in main.tex, keep it:
% \begin{thebibliography}{99}
% ...
% \end{thebibliography}

\end{document}